# Beyond Arabic: Software for Perso-Arabic Script Manipulation

**Alexander Gutkin**[†]  **Cibu Johny**[†]  **Raiomond Doctor**[‡*]  **Brian Roark**[°]  **Richard Sproat**[®]

Google Research

[†]United Kingdom   [‡]India   [°]United States   [®]Japan

{agutkin,cibu,raiomond,roark,rws}@google.com

## Abstract

This paper presents an open-source software library that provides a set of finite-state transducer (FST) components and corresponding utilities for manipulating the writing systems of languages that use the Perso-Arabic script. The operations include various levels of script normalization, including visual invariance-preserving operations that subsume and go beyond the standard Unicode normalization forms, as well as transformations that modify the visual appearance of characters in accordance with the regional orthographies for eleven contemporary languages from diverse language families. The library also provides simple FST-based romanization and transliteration. We additionally attempt to formalize the typology of Perso-Arabic characters by providing one-to-many mappings from Unicode code points to the languages that use them. While our work focuses on the Arabic script diaspora rather than Arabic itself, this approach could be adopted for any language that uses the Arabic script, thus providing a unified framework for treating a script family used by close to a billion people.

## 1 Introduction

While originally developed for recording Arabic, the Perso-Arabic script has gradually become one of the most widely used modern scripts. Throughout history the script was adapted to record many languages from diverse language families, with scores of adaptations still active today. This flexibility is partly due to the core features of the script itself which over the time evolved from a purely consonantal script to include a productive system of diacritics for representing long vowels and optional marking of short vowels and phonological processes such as gemination (Bauer, 1996; Kurzon, 2013). Consequently, many languages productively evolved their own adaptation of the Perso-Arabic script to better suit their phonology by not only augmenting the set of diacritics but also introducing new consonant shapes.

This paper presents an open-source software library designed to deal with the ambiguities and inconsistencies that result from representing various regional Perso-Arabic adaptations in digital media. Some of these issues are due to the Unicode standard itself, where a Perso-Arabic character can often be represented in more than one way (Unicode Consortium, 2021). Others are due to the lack or inadequacies of input methods and the instability of modern orthographies for the languages in question (Aazim et al., 2009; Liljegren, 2018). Such issues percolate through the data available online, such as Wikipedia and Common Crawl (Patel, 2020), negatively impacting the quality of NLP models built with such data. The script normalization software described below goes beyond the standard language-agnostic Unicode approach for Perso-Arabic to help alleviate some of these issues.

The library design is inspired by and consistent with prior work by Johny et al. (2021), introduced in §2, who provided a suite of finite-state grammars for various normalization and (reversible) romanization operations for the Brahmic family of scripts.[1] While the Perso-Arabic script and the respective set of regional orthographies we support – Balochi, Kashmiri, Kurdish (Sorani), Malay (Jawi), Pashto, Persian, Punjabi (Shahmukhi), Sindhi, South Azerbaijani, Urdu and Uyghur – is significantly different from those Brahmic scripts, we pursue a similar finite-state interpretation,[2] as described in §3. Implementation details and simple validation are provided in §4.

---

[*] On contract from Optimum Solutions, Inc.

[1] https://github.com/google-research/nisaba
[2] https://github.com/google-research/nisaba/tree/main/nisaba/scripts/abjad_alphabet

## 2 Related Work

The approach we take in this paper follows in spirit the work of Johny et al. (2021) and Gutkin et al. (2022), who developed a finite-state script normalization framework for Brahmic scripts. We adopt their taxonomy and terminology of low-level script normalization operations, which consist of three types: Unicode-endorsed schemes, such as NFC; further visually-invariant transformations (*visual* normalization); and transformations that modify a character's shape but preserve pronunciation and the overall word identity (*reading* normalization).

The literature on Perso-Arabic script normalization for languages we cover in this paper is scarce. The most relevant work was carried out by Ahmadi (2020) for Kurdish, who provides a detailed analysis of orthographic issues peculiar to Sorani Kurdish along with corresponding open-source script normalization software used in downstream NLP applications, such as neural machine translation (Ahmadi and Masoud, 2020). In the context of machine transliteration and spell checking, Lehal and Saini (2014) included language-agnostic minimal script normalization as a preprocessing step in their open-source $n$-gram-based transliterator from Perso-Arabic to Brahmic scripts. Bhatti et al. (2014) introduced a taxonomy of spelling errors for Sindhi, including an analysis of mistakes due to visually confusable characters. Razak et al. (2018) provide a good overview of confusable characters for Malay Jawi orthography. For other languages the regional writing system ambiguities are sometimes mentioned in passing, but do not constitute the main focus of work, as is the case with Punjabi Shahmukhi (Lehal and Saini, 2012) and Urdu (Humayoun et al., 2022). The specific Perso-Arabic script ambiguities that abound in the online data are often not exhaustively documented, particularly in work focused on multilingual modeling (N. C., 2022; Bapna et al., 2022). As one moves towards lesser-resourced languages, such as Kashmiri and Uyghur, the NLP literature provides no treatment of script normalization issues and the only reliable sources of information are the proposal and discussion documents from the Unicode Technical Committee (e.g., Bashir et al., 2006; Aazim et al., 2009; Pournader, 2014). A forthcoming paper by Doctor et al. (2022) covers the writing system differences between these languages in more detail than we can include in this short paper.

| Op. Type | FST | Language-dep. | Includes |
|---|---|---|---|
| NFC | $\mathcal{N}$ | no | – |
| Common Visual | $\mathcal{V}_c$ | no | $\mathcal{N}$ |
| Visual | $\mathcal{V}$ | yes | $\mathcal{V}_c$ |
| Reading | $\mathcal{R}$ | yes | – |
| Romanization | $\mathcal{M}$ | no | $\mathcal{V}_c$ |
| Transliteration | $\mathcal{T}$ | no | – |

Table 1: Summary of script transformation operations.

One area particularly relevant to this study is the work by the Internet Corporation for Assigned Names and Numbers (ICANN) towards developing a robust set of standards for representing various Internet entities in Perso-Arabic script, such as domain names in URLs. Their particular focus is on *variants*, which are characters that are visually confusable due to identical appearance but different encoding, due to similarity in shape or due to common alternate spellings (ICANN, 2011). In addition, they developed the first proposal to systematize the available Perso-Arabic Unicode code points along the regional lines (ICANN, 2015). These studies are particularly important for cybersecurity (Hussain et al., 2016; Ginsberg and Yu, 2018; Ahmad and Erdodi, 2021), but also inform this work.

This software library is, to the best our knowledge, the first attempt to provide a principled approach to Perso-Arabic script normalization for multiple languages, for downstream NLP applications and beyond.

## 3 Design Methodology

The core components are implemented as individual FSTs that can be efficiently combined together in a single pipeline (Mohri, 2009). These are shown in Table 1 and described below.[3]

**Unicode Normalization** For the Perso-Arabic string encodings which yield visually identical text, the Unicode standard provides procedures that normalize text to a conventionalized normal form, such as the well-known Normalization Form C (NFC), so that visually identical words are mapped to a conventionalized representative of their equivalence class (Whistler, 2021). We implemented the NFC standard as an FST, denoted $\mathcal{N}$ in Table 1, that handles three broad types of transformations: compositions, re-orderings and combinations thereof.

---

[3] When referring to names of Unicode characters we lowercase them and omit the common prefix *arabic* (*letter*).

| FST | Letter | Variant (source) | Canonical |
|---|---|---|---|
| $\mathcal{V}_l^*$ | ⟨ڑ⟩ | reh + small high tah | rreh |
| $\mathcal{V}_l^n$ | ⟨ک⟩ | kaf | keheh |
| $\mathcal{V}_l^f$ | ⟨ى⟩ | alef maksura | farsi yeh |
| $\mathcal{V}_l^i$ | ⟨ه⟩ | heh | heh goal |

Table 2: Example FST components of $\mathcal{V}_l$ for Urdu.

| Op. Type | FST | # states | # arcs | # Kb |
|---|---|---|---|---|
| NFC | $\mathcal{N}$ | 156 | 1 557 | 28.10 |
| Roman. | $\mathcal{M}$ | 32 546 | 52 257 | 1487.10 |
| Translit. | $\mathcal{T}$ | 340 | 518 | 15.15 |

Table 3: Language-agnostic FSTs over UTF-8 strings.

As an example of a first type, consider the *alef with madda above* letter ⟨آ⟩ that can be composed in two ways: as a single character (U+0622) or by adjoining *maddah above* to *alef* ({ U+0627, U+0653 }). The FST $\mathcal{N}$ rewrites the adjoined form into its equivalent composed form. The second type of transformation involves the canonical re-ordering of the Arabic combining marks, for example, the sequence of *shadda* (U+0651) followed by *kasra* (U+0650) is reversed by $\mathcal{N}$. More complex transformations that combine both compositions and re-orderings are possible. For example, the sequence { *alef* (U+0627), *superscript alef* (U+0670), *maddah above* (U+0653) } normalizes to its equivalent form { *alef with madda above* (U+0622), *superscript alef* (U+0670) }.

Crucially, $\mathcal{N}$ is language-agnostic because the NFC standard it implements does not define any transformations that violate the writing system rules of respective languages.

**Visual Normalization** As mentioned in §2, Johny et al. (2021) introduced the term *visual* normalization in the context of Brahmic scripts to denote visually-invariant transformations that fall outside the scope of NFC. We adopt their definition for Perso-Arabic, implementing it as a single language-dependent FST $\mathcal{V}$, shown in Table 1, which is constructed by FST composition: $\mathcal{V} = \mathcal{N} \circ \mathcal{V}_c \circ \mathcal{V}_l$, where ∘ denotes the composition operation (Mohri, 2009).[4]

The first FST after NFC, denoted $\mathcal{V}_c$, is language-agnostic, constructed from a small set of normalizations for visually ambiguous sequences found online that apply to all languages in our library. For example, we map the two-character sequence *waw* (U+0648) followed by *damma* (U+064F) or *small damma* (U+0619) to *u* (U+06C7).

The second set of visually-invariant transformations, denoted $\mathcal{V}_l$, is language-specific and additionally depends on the position within the word. Four special cases are distinguished that are represented as FSTs: position-independent rewrites ($\mathcal{V}_l^*$), isolated-letter rewrites ($\mathcal{V}_l^i$), rewrites in the word-final position ($\mathcal{V}_l^f$), and finally, rewrites in "non-final" word positions, which include visually-identical word-initial and word-medial rewrites ($\mathcal{V}_l^n$). The FST $\mathcal{V}_l$ is composed as $\mathcal{V}_l^i \circ \mathcal{V}_l^f \circ \mathcal{V}_l^n \circ \mathcal{V}_l^*$. Some examples of these transformations for Urdu orthography are shown in Table 2, where the variants shown in the third column are rewritten to their canonical Urdu form in the fourth column.

**Reading Normalization** This type of normalization was introduced for Brahmic scripts by Gutkin et al. (2022), who noted that regional orthographic conventions or lack thereof, which oftentimes conflict with each other, benefit from normalization to some accepted form. Whenever such normalization preserves visual invariance, it falls under the rubric of visual normalization, but other cases belong to *reading* normalization, denoted $\mathcal{R}$ in Table 1. Similar to visual normalization, $\mathcal{R}$ is compiled from language-specific context-dependent rewrite rules. One example of such a rewrite is a mapping from *yeh* ⟨ي⟩ (U+064A) to *farsi yeh* ⟨ى⟩ (U+06CC) in Kashmiri, Persian, Punjabi, Sorani Kurdish and Urdu. For Malay, Sindhi and Uyghur, the inverse transformation is implemented as mandated by the respective orthographies.

For efficiency reasons $\mathcal{R}$ is stored independently of visual normalization $\mathcal{V}$. At run-time, the reading normalization is applied to an input string $s$ as $s' = (s \circ \mathcal{V}) \circ \mathcal{R}$, which is more efficient than $s' = s \circ \mathcal{R}'$, where $\mathcal{R}' = \mathcal{V} \circ \mathcal{R}$.

**Romanization and Transliteration** We also provide language-agnostic romanization ($\mathcal{M}$) and transliteration ($\mathcal{T}$) FSTs. The FST $\mathcal{M}$ converts Perso-Arabic strings to their respective Latin representation in Unicode and is defined as $\mathcal{M} = \mathcal{N} \circ \mathcal{V}_c \circ \mathcal{M}_c$, where $\mathcal{N}$ and $\mathcal{V}_c$ were described above, and $\mathcal{M}_c$ implements a one-to-one mapping from 198 Perso-Arabic characters to their respective romanizations using our custom romanization scheme derived from language-specific Library of Congress rules (LC, 2022) and various ISO standards (ISO, 1984, 1993, 1999). For example, in our scheme the Uyghur *yu* ⟨ۈ⟩ (U+06C8) maps to ⟨ü⟩. The transliteration FST $\mathcal{T}$ converts the strings from Unicode Latin into Perso-Arabic. It is smaller than $\mathcal{M}$ and is defined as $\mathcal{T} = \mathcal{M}_c^{-1}$.

---
[4] See Johny et al. (2021) for details on FST composition and other operations used in this kind of script normalization.

| Language Information | | Visual Normalization ($\mathcal{V}$) | | | Reading Normalization ($\mathcal{R}$) | | |
|---|---|---|---|---|---|---|---|
| Code | Name | # states | # arcs | # Mb | # states | # arcs | # Mb |
| azb | South Azerbaijani | 315 933 | 635 647 | 16.49 | 21 | 735 | 0.012 |
| bal | Balochi | 620 226 | 1 244 472 | 32.31 | 24 | 738 | 0.013 |
| ckb | Kurdish (Sorani) | 1 097 937 | 2 199 732 | 57.15 | 39 | 753 | 0.013 |
| fa | Persian | 940 436 | 1 884 347 | 48.96 | 36 | 750 | 0.013 |
| ks | Kashmiri | 1 772 494 | 3 547 448 | 92.21 | 44 | 794 | 0.014 |
| ms | Malay | 199 777 | 403 373 | 10.45 | 21 | 735 | 0.012 |
| pa | Punjabi | 2 050 154 | 4 105 465 | 106.69 | 24 | 738 | 0.013 |
| ps | Pashto | 291 564 | 587 552 | 15.23 | 24 | 738 | 0.013 |
| sd | Sindhi | 1 703 726 | 3 403 283 | 88.53 | 34 | 748 | 0.013 |
| ug | Uyghur | 1 255 054 | 2 513 231 | 65.31 | 24 | 738 | 0.013 |
| ur | Urdu | 2 071 139 | 4 138 950 | 107.65 | 31 | 745 | 0.013 |

Table 4: Summary of FSTs over UTF-8 strings for visual and reading normalization.

**Character-Language Mapping** The geography and scope of Perso-Arabic script adaptations is vast. To document the typology of the characters we developed an easy-to-parse mapping between the characters and the respective languages and/or macroareas that relate to a group of languages building on prior work by ICANN (2015). For example, using this mapping it is easy to find that the letter *beh with small v below* ⟨ࢠ⟩ (U+08A0) is part of the orthography of Wolof, a language of Senegal (Ngom, 2010), while *gaf with ring* ⟨ڰ⟩ (U+06B0) belongs to Saraiki language spoken in Pakistan (Bashir and Conners, 2019). This mapping can be used to auto-generate the orthographic inventories for lesser-resourced languages.

## 4 Software Details and Validation

Our software library is implemented using Pynini, a Python library for constructing finite-state grammars and for performing operations on FSTs (Gorman, 2016; Gorman and Sproat, 2021). Each FST is compiled from the collections of individual context-dependent letter rewrite rules (Mohri and Sproat, 1996) and is available in two versions: over an alphabet of UTF-8 encoded bytes and over the integer Unicode code points. The FSTs are stored uncompressed in binary FST archives (FARs) in OpenFst format (Allauzen et al., 2007).

The summaries of language-agnostic and language-dependent FSTs over UTF-8 strings are shown in Table 3 and Table 4, respectively. As can be seen from the tables, the language-agnostic and reading normalization FSTs are relatively uncomplicated and small in terms of number of states, arcs and the overall (uncompressed) size on disk. The visual normalization FSTs are significantly larger, which is explained by the number of composition operations used in their construction (see §3). The reading normalization FSTs for South Azerbaijani and Malay shown in Table 4 implement the identity mapping. This is because we could not find enough examples requiring reading-style normalization in online data (see the Limitations section for more details).

| Lang. | $s' = s \circ \mathcal{V}$ | | $s' = (s \circ \mathcal{V}) \circ \mathcal{R}$ | |
|---|---|---|---|---|
| | % tokens | % types | % tokens | % types |
| ckb | 18.27 | 25.84 | 30.07 | 41.26 |
| sd | 17.32 | 14.83 | 21.74 | 17.31 |
| ur | 0.09 | 1.16 | 0.10 | 1.23 |

Table 5: Percentage of tokens and types changed.

As an informal sanity check we validate the prevalence of normalization on word-frequency lists for Sorani Kurdish (ckb), Sindhi (sd) and Uyghur (ug) from project Crúbadán (Scannell, 2007). Table 5 shows the percentages of tokens and types changed ($s' \neq s$) by visual normalization on one hand and the combined visual and reading normalization on the other. Urdu has the fewest number of modifications compared to Sorani Kurdish and Sindhi, most likely due to a more regular orthography and stable input methods manifest in the crawled data. Significantly more extensive analysis and experiments in statistical language modeling and neural machine translation for the languages covered in this paper are presented in a forthcoming study (Doctor et al., 2022).

**Example** The use of the library is demonstrated by the following Python example that implements a simple command-line utility for performing reading normalization on a single string using Pynini APIs. The program requires two FAR files that store compiled visual and reading normalization grammars, the upper-case BCP-47 language code for retrieving the FST for a given language, and an input string:[5]

| Lang. | Input | Output | Correct Output |
|---|---|---|---|
| bal | دئیٹ | دئیٹ | *teh* |
| ckb | لەشکر | لەشکر | *keheh* |
| fa | مؤسسه | موسسه | *waw* |
| ks | ہےتک | ہیتک | *kashmiri yeh* |
| pa | کئی | کئی | *farsi yeh* |
| sd | ڳوهه | ڳوہہ | *heh goal* |
| ug | ساى | ساي | *yeh* |
| ur | صورة | صورۃ | *teh marbuta goal* |

Table 6: Some examples of reading normalization.

```
example.py
from absl import app
from absl import flags
from collections.abc import Iterable, Sequence
import pynini as pyn

flags.DEFINE_string("input", None, "Input string.")
flags.DEFINE_string("lang", None, "Language code.")
flags.DEFINE_string("reading_grm", None, "Reading FAR.")
flags.DEFINE_string("visual_grm", None, "Visual FAR.")
FLAGS = flags.FLAGS

def load_fst(grammar_path: str, lang: str) -> pyn.Fst:
  """Loads FST for specified grammar and language."""
  return pyn.Far(grammar_path)[lang]

def apply(text: str, fsts: Iterable[pyn.Fst]) -> str:
  """Applies sequence of FSTs on an input string."""
  try:
    composed = pyn.escape(text)
    for fst in fsts:
      composed = (composed @ fst).optimize()
    return pyn.shortestpath(composed).string()
  except pyn.FstOpError as error:
    raise ValueError(f"Error for string `{text}`")

def main(argv: Sequence[str]) -> None:
  # ... initializing FLAGS
  visual_fst = load_fst(FLAGS.visual_grm, FLAGS.lang)
  reading_fst = load_fst(FLAGS.reading_grm, FLAGS.lang)
  out = apply(FLAGS.input, [visual_fst, reading_fst])
  print(f"=> {out}")

if __name__ == "__main__":
  app.run(main)
```

The visual and reading FSTs for a given language are retrieved from the relevant FAR files using `load_fst` function. The input string is first converted to a linear FST. The visual and reading normalization FSTs are then sequentially composed with the input FST and a shortest path algorithm is applied on the result, which is then converted from a linear FST back to a Python string in `apply` function to yield the final normalized output.

Some examples of reading normalization produced using the `example.py` utility above for some of the supported languages are shown in Table 6. For each language, the input string in the second column of the table is normalized to a string shown in the third column. The final column shows the name of a particular letter in the output string that replaced the original letter from the input string, e.g., for Sorani Kurdish (ckb) the following rewrite occurs: *swash kaf* (U+06AA) → *keheh* (U+06A9), while for Punjabi (pa), *yeh* (U+064A) → *farsi yeh* (U+06CC).

## 5 Conclusion and Future Work

We have presented a flexible FST-based software package for low-level processing of orthographies based on Perso-Arabic script. We described the main components of the architecture consisting of various script normalization operations, romanization/transliteration, and character-language index. We expect to increase the current language coverage of eleven languages to further relatively well-documented orthographies, but also provide treatment for resource-scarce orthographies, such as the Ajami orthographies of Sub-Saharan Africa (Mumin, 2014).

## Limitations

When developing the visual and reading normalization rules for the eleven languages described in this paper we made use of publicly available online data consisting of the respective Wikipedias, Wikipron (Lee et al., 2020), Crúbadán (Scannell, 2007) and parts of Common Crawl (Patel, 2020). The latter corpus is particularly noisy and requires non-trivial filtering (Kreutzer et al., 2022). Furthermore, many Wikipedia and Common Crawl documents contain code-switched text in several languages that are recorded in Perso-Arabic. Robust language identification (LID) is required to distinguish between tokens in such sentences (for example, Kashmiri vs. Pashto or Balochi) in order not to confuse between the respective orthographies. Since we did not have access to robust LID models for the languages under study, for lesser-resourced languages such as Kashmiri, Malay in Jawi orthography, South Azerbaijani and Uyghur, it is likely that some of the words we used as examples requiring normalization may have been misclassified resulting in normalizations that should not be there.

---
[5]The infrastructure for compiling the Pynini grammars is described in Johny et al. (2021).